\documentclass[conference]{IEEEtran}
\IEEEoverridecommandlockouts
\usepackage{cite}
\usepackage{amsmath,amssymb,amsfonts}
\usepackage{algorithmic}
\usepackage{graphicx}
\usepackage{textcomp}
\usepackage{xcolor}
\usepackage{url}
\def\BibTeX{{\rm B\kern-.05em{\sc i\kern-.025em b}\kern-.08em
    T\kern-.1667em\lower.7ex\hbox{E}\kern-.125emX}}
\begin{document}

\title{BROTHER: Behavioral Recognition Optimized Through Heterogeneous Ensemble Regularization for Ambivalence and Hesitancy}

\author{
    \IEEEauthorblockN{1\textsuperscript{st} Alexandre Pereira}
    \IEEEauthorblockA{
    \textit{University of Pernambuco}\\
    Recife, Brazil \\
    armp@ecomp.poli.br}
    \and
    \IEEEauthorblockN{2\textsuperscript{nd} Bruno Fernandes}
    \IEEEauthorblockA{
    \textit{University of Pernambuco}\\
    Recife, Brazil \\
    bjtf@ecomp.poli.br}
    \and
    \IEEEauthorblockN{3\textsuperscript{rd} Pablo Barros}
    \IEEEauthorblockA{
    \textit{University of Pernambuco}\\
    Recife, Brazil \\
    pablovin@gmail.com}
}

\maketitle

\begin{abstract}
Recognizing complex behavioral states such as Ambivalence and Hesitancy (A/H) in naturalistic video settings remains a significant challenge in affective computing. Unlike basic facial expressions, A/H manifests as subtle, multimodal conflicts that require deep contextual and temporal understanding. In this paper, we propose a highly regularized, multimodal fusion pipeline to predict A/H at the video level. We extract robust unimodal features from visual, acoustic, and linguistic data, introducing a specialized statistical text modality explicitly designed to capture temporal speech variations and behavioral cues. To identify the most effective representations, we evaluate 15 distinct modality combinations across a committee of machine learning classifiers (MLP, Random Forest, and GBDT), selecting the most well-calibrated models based on validation Binary Cross-Entropy (BCE) loss. Furthermore, to optimally fuse these heterogeneous models without overfitting to the training distribution, we implement a Particle Swarm Optimization (PSO) hard-voting ensemble. The PSO fitness function dynamically incorporates a train-validation gap penalty ($\lambda$) to actively suppress redundant or overfitted classifiers. Our comprehensive evaluation demonstrates that while linguistic features serve as the strongest independent predictor of A/H, our heavily regularized PSO ensemble ($\lambda = 0.2$) effectively harnesses multimodal synergies, achieving a peak Macro F1-score of 0.7465 on the unseen test set. These results emphasize that treating ambivalence and hesitancy as a multimodal conflict, evaluated through an intelligently weighted committee, provides a robust framework for in-the-wild behavioral analysis.
\end{abstract}


\section{Introduction}

Ambivalence and hesitancy (A/H) are closely related psychological states that often act as the primary barriers when people try to start, maintain, or successfully change health-related behaviors \cite{gonzalez2026bah}. Unlike clear-cut emotions such as anger or joy, A/H manifests as a subtle inner conflict. It places an individual in a "gray area" between positive and negative perspectives. Because these behavioral cues are often quiet and vary significantly from person to person, detecting them automatically requires a careful analysis of visual expressions, vocal tone, and word choice over time.

The dataset utilized for this study is derived from the 10th Affective Behavior Analysis in-the-wild (ABAW) competition, specifically the Ambivalence and Hesitancy Video Recognition Challenge \cite{abaw10_ah_challenge}. The Behavioural Ambivalence/Hesitancy (BAH) dataset features a diverse group of participants recorded in natural, everyday settings, providing a realistic foundation for analyzing complex human behaviors outside of a controlled environment.

Standard emotion recognition systems often struggle with A/H because they try to force every input into a fixed category, like "happy" or "sad". This approach ignores the contradictions and shifts in timing that are actually the indicators of a hesitant or ambivalent state. To solve this, we propose a flexible multimodal pipeline designed to capture these deeper patterns in how people speak, look, and sound.

Our approach includes three powerful state-of-the-art models to translate raw data into meaningful information: SigLip2 for visual frames \cite{siglip2}, HuBERT for audio \cite{hubert}, and F2LLM for text transcripts \cite{f2llm}. Recognizing that A/H is often revealed through speech pacing and specific phrasing, we also created a fourth modality consisting entirely of statistical features from the other three modalities. It summarizes how the data changes over time and uses targeted strategies to measure localized hesitancy and broader ambivalence.

Instead of relying on a single, rigid AI model, we built a "committee" of different models. We trained three different types of machine learning architectures: Multi-Layer Perceptrons, Random Forests, and Gradient Boosted Decision Trees, across every possible combination of our four data sources. For each combination, we kept only the most accurate and well-calibrated model. To combine the final 15 predictions, we used Particle Swarm Optimization (PSO) \cite{pso} to find the best voting weights. This process was designed to ensure the system actually learned to generalize to new people.

By allowing the system to discover complex relationships between different modalities without forcing them into basic emotional boxes, our pipeline proved to be highly effective. In our tests, this methodology achieved a validation Macro F1-score of 0.7578 and a test Macro F1-score of 0.7465, demonstrating that combining specialized feature extraction with a smart voting committee is a powerful way to understand complex human behavior.

\section{Related Work}

Gonz{\'a}lez-Gonz{\'a}lez et al. \cite{gonzalez2026bah} introduced the BAH dataset to study ambivalence and hesitancy in naturalistic settings. To establish a baseline, they employed a zero-shot prompting strategy using the Video-LLaVA Multimodal Large Language Model (M-LLM). By prompting the model with sampled video frames alongside the video's transcript, they achieved a video-level macro F1-score of 0.6341. This demonstrated the feasibility of using generalized vision-language architectures for behavioral prediction without requiring task-specific temporal training.

In contrast, the HSEmotion team \cite{hsemotion} proposed an efficient multimodal fusion pipeline. They extracted frame-level facial emotional descriptors using their custom EmotiEffLib library, combining them with acoustic features and textual speech embeddings. These localized features were temporally aggregated and processed through lightweight classifiers, such as a Multi-Layer Perceptron (MLP). This approach highlighted the efficacy of fusing robust unimodal feature extractors to model complex affective states, providing a computationally lighter alternative to generative models.

\section{Methodology}

\subsection{Dataset Context and Modality Definition}

The dataset utilized for this study is derived from the 10th Affective Behavior Analysis in-the-wild (ABAW) competition, specifically targeting the Ambivalence and Hesitancy challenge \cite{abaw10_ah_challenge}. This Behavioural Ambivalence/Hesitancy (BAH) dataset features a diverse array of participants recorded in naturalistic settings, providing a rich foundation for analyzing complex human behaviors \cite{gonzalez2026bah}. 

To capture a comprehensive and multidimensional representation of participant behavior, features were extracted across four distinct modalities. The first three modalities represent the direct data streams: Visual (video frames), Textual (transcripts), and Audio (vocal acoustics). 

The fourth modality is entirely composed of Statistical Features, which are mathematically derived from the localized chunks of the first three modalities. This statistical modality serves to aggregate temporal and structural patterns that might otherwise be lost in direct sequential processing.

We explicitly avoided pre-trained classification models that reduce outputs to fixed or basic emotions. Relying on such rigid architectures could hinder the classification pipeline by preventing more complex, nuanced, and multimodal feature relationships from emerging organically during the training process.

\begin{figure}[htbp]
\centerline{\includegraphics[width=\columnwidth]{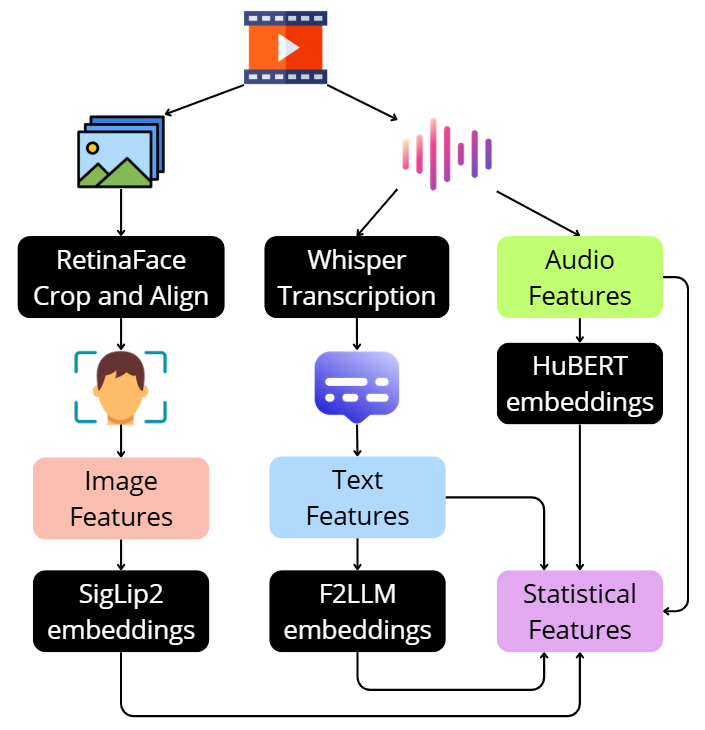}}
\caption{High-level overview of the multimodal feature extraction pipeline, illustrating the parallel processing of Visual, Audio, Textual, and Statistical modalities.}
\label{fig:pipeline}
\end{figure}

\subsubsection{Visual Extraction with RetinaFace and SigLip2}

For the visual modality, the participant's face is first detected, cropped, and aligned for each frame using RetinaFace. Each preprocessed face image then has its features extracted using SigLip2 \cite{siglip2}. 

This foundational model was chosen because it represents a significant improvement upon its predecessor, SigLip \cite{siglip}, which is the highest-ranking available model on the Massive Image Embedding Benchmark (MIEB) for zero-shot image classification \cite{mieb}. Aligning the image latent space to a textual latent space alleviates visual space complexity by mapping raw pixels to descriptive semantic regions, while also providing stronger cultural and fairness guarantees.

After the visual feature extraction, the embeddings' magnitudes are normalized. The average cosine similarity of the embeddings throughout the entire video is computed and passed through a Median Absolute Deviation (MAD) filter using a threshold multiplier of 50. This filtering process effectively removes noisy frames where the camera was adjusting to illumination changes or where no face was visibly present.

To preserve temporal dynamics that standard statistical pooling might miss \cite{statspool}, the filtered frames are passed through three averaging functions, creating a compact representation of 768 components each: the normal features, the first derivative, and the second derivative. These three averaged features are concatenated to yield 2304 dimensions. 

Finally, this visual feature vector is scaled using a Standard Scaler and reduced via Principal Component Analysis (PCA) down to 512 dimensions, safely retaining 99\% of the original explained variance. It is important to note that PCA and Standard Scaler normalization were exclusively applied to the video embeddings and the statistical modality, rather than the text or audio embeddings.

\subsubsection{Audio Extraction with HuBERT}

HuBERT \cite{hubert} was chosen for audio extraction due to its proven success in previous editions of the ABAW competition and its highly effective unsupervised training paradigm. 

This architectural design makes it ideal for capturing non-linguistic audio cues, such as tone and rhythm, without imposing predefined emotional categories onto the acoustic data. Audio features were extracted in one-second chunks, mirroring the visual extraction timeline, with their embedding magnitudes subsequently normalized to maintain scale consistency.

\subsubsection{Textual Extraction with F2LLM}

F2LLM \cite{f2llm} was selected for textual embeddings due to its leading classification ranking on the Massive Text Embedding Benchmark (MTEB) \cite{mteb}. Because the model was trained exclusively on open-source, non-synthetic data, it demonstrated a superior capacity to capture the authentic nuances and colloquialisms of human speech present in the ABAW dataset. 

Embeddings were extracted for the entire video transcript at once to provide a rich global context, and their magnitudes were normalized prior to further processing. 

\subsection{The Fourth Modality: Statistical Feature Extraction}

The fourth modality serves as a structured aggregation of the temporal data. Second by second (video and audio) and sentence-level (text) sequences from the primary modalities were collapsed using statistical pooling, which computes the minimum, maximum, mean, and standard deviation across the sequence. Global features were represented as single continuous values. All resulting statistical features were subsequently normalized using a Standard Scaler.

\subsubsection{Visual and Audio Statistics}

Visual statistics were calculated to summarize the quality and consistency of the one-second video chunks. The visual statistical features include:
\begin{itemize}
    \item \textbf{Valid Chunk:} A binary indicator marking whether the one-second chunk contains valid frames that survived the MAD filter.
    \item \textbf{Valid Ratio:} The proportion of valid frames within that specific chunk.
    \item \textbf{Similarity Mean:} The mean cosine similarity of the frames within the chunk, indicating visual stability.
\end{itemize}

Simultaneously, acoustic characteristics were extracted for each one-second audio chunk using the \texttt{librosa} library. The audio statistical features include:
\begin{itemize}
    \item \textbf{RMS:} Root Mean Square energy, representing the amplitude and loudness.
    \item \textbf{Spectral Centroid and Bandwidth:} The center of mass of the spectrum indicating the "brightness" of the audio, and the width of the spectral band.
    \item \textbf{Spectral Rolloff:} The frequency below which a specified percentage of the total spectral energy lies.
    \item \textbf{Zero-Crossing Rate:} The rate at which the signal changes sign, which is useful for detecting percussive or erratic speech sounds.
    \item \textbf{Silence Ratio:} The proportion of the chunk where the amplitude falls below a -30 dB threshold.
    \item \textbf{Pitch Mean and Standard Deviation:} The average fundamental frequency (f0) computed via the YIN algorithm, along with its variance inside the chunk.
\end{itemize}

\subsubsection{Textual Statistics and Behavioral Language Strategies}

Textual statistics were heavily enriched by analyzing conversational behaviors, specifically focusing on hesitant and ambivalent language patterns. 

Standard textual statistics included global word counts, frequencies of short pauses (commas) and long pauses (terminal punctuation), consecutive word repetitions, and overall lexical diversity (the ratio of unique words to total words).

Beyond standard counts, we implemented highly specialized strategies to detect behavioral markers, which are inspired on the language cues defined at \cite{gonzalez2026bah}:
\begin{itemize}
    \item \textbf{Hesitant Language:} Hesitancy is highly localized and was therefore computed strictly on a per-sentence basis. Each sentence was compared against predefined dictionaries containing 10 to 20 representative words or short expressions for four hesitancy categories: filler words, filler sounds, hedging, and corrections. The algorithm calculated the cosine similarity between the sentence embedding and each expression. The system could then determine how much larger one similarity score was relative to the others, accurately isolating moments of localized hesitation.
    
    \item \textbf{Ambivalent Language:} Unlike hesitancy, ambivalence requires an understanding of broader conversational context and was therefore computed over the whole text. The extraction relied on a structured prompt-based embedding strategy. Specifically, the prompt was structured using conversational tokens and populated with 10 representative example sentences for each category (sentiment, capability, excuse, success, motivation and opportunity). To strictly target the ambivalent nature of the text, we conceptualized four distinct sentence poles: neutral, negative, positive, and both. The target text embeddings were compared to the embedded prompts, and a softmax function was applied to the resulting cosine similarities with a temperature multiplier to accentuate the differences between these poles.
\end{itemize}

\subsection{Multimodal Classification Committee}

Given the four distinct feature modalities (Text, Audio, Video, and Stats), there are 15 possible combinations of modalities ($2^4 - 1$). Rather than relying on a single generalized algorithm to handle all combinations, we implemented a heterogeneous classification committee to ensure the strongest possible representation for each specific subset of data. 

For each of the 15 modality combinations, three distinct machine learning architectures were trained:
\begin{enumerate}
    \item \textbf{Multi-Layer Perceptron (MLP):} A deep neural network with three hidden layers featuring Gaussian noise injection, Batch Normalization, Dropout, and RELU activations.
    \item \textbf{Random Forest (RF):} Configured via LightGBM with balanced class weights and a depth limit of 50.
    \item \textbf{Gradient Boosted Decision Trees (GBDT):} Configured via LightGBM with a low learning rate ($1e-3$) for robust sequential boosting.
\end{enumerate}

Once trained, the candidate models were evaluated on the validation set using Binary Cross-Entropy (BCE) loss. BCE was chosen over hard-classification metrics because it strictly evaluates the calibration and confidence of the predicted probabilities rather than arbitrary threshold boundaries. 

The algorithm that achieved the lowest validation BCE loss was selected as the permanent representative for that modality combination, discarding the other two. This methodology resulted in a diverse final ensemble of 15 highly optimized base models. Following the selection of the best algorithm for each combination, an optimal binary classification threshold was computed based on the validation set F1-score.

\subsection{Particle Swarm Optimization (PSO) for Hard Voting}

To intelligently fuse the predictions of the 15 committee members, we utilized Particle Swarm Optimization (PSO) to discover the optimal voting weights. The PSO navigated the 15-dimensional continuous weight space to maximize ensemble performance while heavily penalizing overfitting. 

Unlike soft-voting approaches that average probability distributions, our ensemble utilized a hard-voting mechanism. Each model cast a binary vote based on its optimized threshold, and the final classification was determined by checking if the weighted sum of these binary votes exceeded half of the total weight pool.

The PSO was initialized with 50 particles to ensure broad spatial coverage, operating over 100 epochs to allow the regularization penalties to stabilize. The swarm dynamics were controlled by an inertia weight ($w = 0.9$) to encourage wide exploration, a cognitive parameter ($c_1 = 1.5$) allowing particles to respect their historical personal bests, and a social parameter ($c_2 = 2.1$) guiding them toward the global best.

Crucially, to prevent the ensemble from simply memorizing the training data, the fitness function was designed to maximize the Harmonic Mean of the Training F1 and Validation F1 scores, mathematically forcing the models to generalize. A squared penalty was deducted based on the absolute gap between the train and validation scores, controlled by a penalty coefficient ($\lambda$):

\begin{equation}
Fitness = \frac{2 \cdot F1_{val} \cdot F1_{train}}{F1_{val} + F1_{train}} - (\lambda \cdot |F1_{train} - F1_{val}|)^2
\label{eq:fitness}
\end{equation}

To empirically determine the best regularization balance, the entire PSO process was executed across five independent runs, varying the $\lambda$ penalty weight from 0.0 to 0.8 in increments of 0.2. The weights resulting from those independent runs were extracted from the best iteration and sent as the five allowed trials for the competition.

\section{Results and Discussion}

\subsection{Committee Model Selection}

The first phase of our evaluation focused on establishing the strongest candidate model (MLP, RF, or GBDT) for each of the 15 modality combinations. Models were selected strictly based on achieving the lowest Binary Cross-Entropy (BCE) loss on the validation set, ensuring that the selected models provided well-calibrated confidence scores rather than merely fitting a hard classification boundary.

Table \ref{tab:committee} details the validation BCE loss and Macro F1-scores for all candidate models across all modality combinations. 

\begin{table}[htbp]
\caption{Validation BCE Loss and Macro F1-Score for Committee Candidates}
\begin{center}
\resizebox{\columnwidth}{!}{%
\begin{tabular}{|l|cc|cc|cc|c|}
\hline
\textbf{Modality} & \multicolumn{2}{c|}{\textbf{MLP}} & \multicolumn{2}{c|}{\textbf{RF}} & \multicolumn{2}{c|}{\textbf{GBDT}} & \textbf{Winner} \\
\textbf{Combination} & BCE & F1 & BCE & F1 & BCE & F1 & (Best BCE) \\
\hline
Text & \textbf{0.5730} & 0.7275 & 0.6234 & 0.6613 & 0.6309 & 0.6779 & MLP \\
Audio & \textbf{0.6751} & 0.6317 & 0.6950 & 0.5992 & 0.6922 & 0.5973 & MLP \\
Video & 0.7465 & 0.5226 & 0.6963 & 0.4638 & \textbf{0.6956} & 0.4704 & GBDT \\
Stats & 0.6496 & 0.6928 & \textbf{0.6341} & 0.6406 & 0.6403 & 0.6317 & RF \\
\hline
Text+Audio & \textbf{0.5925} & 0.7014 & 0.6322 & 0.6409 & 0.6393 & 0.6544 & MLP \\
Text+Video & 0.6884 & 0.5363 & \textbf{0.6242} & 0.6686 & 0.6316 & 0.6693 & RF \\
Text+Stats & \textbf{0.5937} & 0.6512 & 0.6203 & 0.6935 & 0.6294 & 0.6773 & MLP \\
Audio+Video & 0.7170 & 0.4959 & 0.6936 & 0.5989 & \textbf{0.6921} & 0.5921 & GBDT \\
Audio+Stats & \textbf{0.6612} & 0.6282 & 0.6698 & 0.6261 & 0.6679 & 0.6232 & MLP \\
Video+Stats & 0.7283 & 0.6228 & \textbf{0.6593} & 0.6521 & 0.6621 & 0.6494 & RF \\
\hline
Text+Aud+Vid & 0.6872 & 0.5008 & \textbf{0.6349} & 0.6544 & 0.6414 & 0.6529 & RF \\
Text+Aud+Sts & \textbf{0.6001} & 0.6314 & 0.6282 & 0.6601 & 0.6370 & 0.6574 & MLP \\
Text+Vid+Sts & 0.7334 & 0.5426 & \textbf{0.6209} & 0.6838 & 0.6302 & 0.6744 & RF \\
Aud+Vid+Sts & 0.6897 & 0.5377 & \textbf{0.6732} & 0.6288 & 0.6736 & 0.6220 & RF \\
\hline
All Modalities & 0.6962 & 0.5951 & \textbf{0.6271} & 0.6601 & 0.6362 & 0.6540 & RF \\
\hline
\end{tabular}
}
\label{tab:committee}
\end{center}
\end{table}

The unimodal results indicate that the Text modality is the strongest independent predictor of ambivalence and hesitancy (Validation F1 of 0.7275 via MLP), highlighting the profound importance of transcript-based contextual language. Conversely, the standalone Video modality struggled (Validation F1 of 0.4704), underscoring the difficulty of detecting subtle inner conflict using facial features alone without temporal or acoustic context. 

Interestingly, while the MLP architecture dominated the simpler configurations, the Random Forest (RF) algorithm consistently won the highly dimensional, complex multimodal combinations (winning 7 out of 15, including the combination of all four modalities). This demonstrates the RF's superior ability to handle heterogeneous feature spaces and avoid overfitting when presented with concatenated multimodal embeddings.

\subsection{Particle Swarm Optimization (PSO) Ensemble F1 Optimization}

After establishing the 15-model committee, the PSO algorithm was utilized to find the optimal hard-voting weights. To thoroughly test the robustness of our anti-overfitting strategy, five separate PSO runs were conducted, systematically increasing the train-validation gap penalty ($\lambda$) from 0\% to 80\%.

Table \ref{tab:pso_macro} presents the resulting Macro F1-scores, while Table \ref{tab:pso_weighted} displays the Weighted F1-scores across these different penalty configurations. 

\begin{table}[htbp]
\caption{PSO Ensemble Performance: Macro F1-Scores}
\begin{center}
\begin{tabular}{|c|c|c|c|}
\hline
\textbf{Penalty ($\lambda$)} & \textbf{Train F1-M} & \textbf{Val F1-M} & \textbf{Test F1-M} \\
\hline
0.0 (0\%) & 0.9743 & 0.7355 & 0.7399 \\
0.2 (20\%) & \textbf{0.9820} & 0.7355 & \textbf{0.7465} \\
0.4 (40\%) & 0.9653 & \textbf{0.7578} & 0.7409 \\
0.6 (60\%) & 0.9653 & \textbf{0.7578} & 0.7409 \\
0.8 (80\%) & 0.9781 & 0.7486 & 0.7424 \\
\hline
\end{tabular}
\label{tab:pso_macro}
\end{center}
\end{table}

\begin{table}[htbp]
\caption{PSO Ensemble Performance: Weighted F1-Scores}
\begin{center}
\begin{tabular}{|c|c|c|c|}
\hline
\textbf{Penalty ($\lambda$)} & \textbf{Train F1-W} & \textbf{Val F1-W} & \textbf{Test F1-W} \\
\hline
0.0 (0\%) & 0.9743 & 0.7485 & 0.7510 \\
0.2 (20\%) & \textbf{0.9820} & 0.7485 & \textbf{0.7559} \\
0.4 (40\%) & 0.9653 & \textbf{0.7672} & 0.7491 \\
0.6 (60\%) & 0.9653 & \textbf{0.7672} & 0.7491 \\
0.8 (80\%) & 0.9781 & 0.7588 & 0.7508 \\
\hline
\end{tabular}
\label{tab:pso_weighted}
\end{center}
\end{table}

As illustrated in Tables \ref{tab:pso_macro} and \ref{tab:pso_weighted}, incorporating the regularization penalty yielded tangible improvements in generalization. A penalty of 40\% and 60\% produced the highest Validation Macro and Weighted F1-scores, while the 20\% penalty achieved the absolute highest metrics on the unseen Test set (0.7465 Macro F1 and 0.7559 Weighted F1). 

\begin{figure}[htbp]
\centerline{\includegraphics[width=0.8\columnwidth]{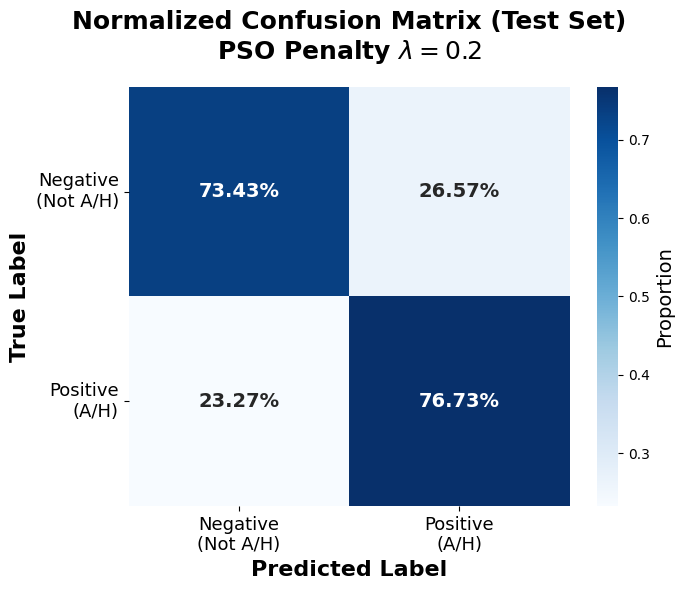}}
\caption{Normalized Confusion Matrix for the best performing ensemble on the Test Set (PSO $\lambda = 0.2$), achieving a Test Macro F1 of 0.7465.}
\label{fig:cm_best}
\end{figure}

An analysis of the final voting weights generated by the PSO reveals its intrinsic capacity for feature selection. Under a 0\% penalty, the algorithm distributed positive weights across 12 of the 15 models. However, as the penalty increased to 80\%, the PSO became highly selective, completely zeroing out 9 models and concentrating the voting power on only the 6 most reliable multimodal configurations. The Text modality, along with the Text+Video+Stats combination, consistently retained the highest weights across all penalty iterations, proving the PSO actively suppressed redundant classifiers to optimize the ensemble's collective reasoning. Figure \ref{fig:cm_best} visualizes the normalized confusion matrix for our top-performing test configuration ($\lambda=0.2$).

\section{Conclusion}

In this paper, we introduced a robust multimodal pipeline for the automatic detection of ambivalence and hesitancy in naturalistic video settings. Recognizing that A/H is not a static emotion but a complex state of behavioral conflict, our approach deliberately bypassed rigid, pre-trained emotional classifiers. 

Instead, we leveraged foundational unimodal architectures (SigLip2, HuBERT, and F2LLM) alongside a uniquely engineered statistical modality. This fourth modality was explicitly designed to capture temporal variations and target specialized behavioral language concepts, such as contextual ambivalence and localized hesitancy, directly from the extracted transcripts.

By distributing these modalities across a heterogeneous committee of optimized machine learning classifiers (MLP, RF, and GBDT), we ensured that the unique strengths of each data subset were mathematically preserved. Furthermore, the integration of a Particle Swarm Optimization hard-voting ensemble, driven by a fitness function that actively penalizes the training-validation gap, effectively prevented the pipeline from memorizing noise and heavily suppressed redundant models. 

Our empirical results demonstrate that while linguistic features serve as the primary anchor for detecting A/H, achieving peak generalization requires the intelligent, dynamically weighted fusion of visual and acoustic signals. Ultimately, treating ambivalence and hesitancy as a multimodal temporal conflict, evaluated by a heavily regularized committee, yields highly reliable recognition capabilities.

\vspace{12pt}

\end{document}